\documentclass[conference]{IEEEtran}
\IEEEoverridecommandlockouts
\usepackage{graphicx}
\usepackage{amsmath} 
\usepackage{bbm}
\ifCLASSINFOpdf
\else
\fi

\usepackage{algorithm2e}
\SetKwComment{Comment}{/* }{ */}
\RestyleAlgo{ruled}
\LinesNumbered

\usepackage{svg}
\usepackage{multirow}

\begin{document}

\title{On the Definition of Robustness and Resilience of AI Agents for Real-time Congestion Management\\
\thanks{The research leading to this work is part of the AI4REALNET (\textit{AI for REAL-world NETwork operation}) project, which received funding from European Union’s Horizon Europe Research and Innovation programme under the Grant Agreement No 101119527, and from the Swiss State Secretariat for Education, Research and Innovation (SERI). This project is funded by the European Union and SERI. Views and opinions expressed are however those of the author(s) only and do not necessarily reflect those of the European Union and SERI. Neither the European Union nor the granting authority can be held responsible for them.}
}

\author{\IEEEauthorblockN{Timothy Tjhay, Ricardo J. Bessa, José Paulos}
\IEEEauthorblockA{Center for Power and Energy Systems\\
INESC TEC\\
Porto, Portugal\\
\{timothy.tjhay; ricardo.j.bessa; jose.paulos\}@inesctec.pt}
}

\maketitle

\begin{abstract}
The European Union's Artificial Intelligence (AI) Act defines robustness, resilience, and security requirements for high-risk sectors but lacks detailed methodologies for assessment. This paper introduces a novel framework for quantitatively evaluating the robustness and resilience of reinforcement learning agents in congestion management. Using the AI-friendly digital environment Grid2Op, perturbation agents simulate natural and adversarial disruptions by perturbing the input of AI systems without altering the actual state of the environment, enabling the assessment of AI performance under various scenarios. Robustness is measured through stability and reward impact metrics, while resilience quantifies recovery from performance degradation. The results demonstrate the framework’s effectiveness in identifying vulnerabilities and improving AI robustness and resilience for critical applications.

\end{abstract}


\IEEEpeerreviewmaketitle

\section{Introduction}


The real-time operation of power grids requires efficient power flow management, especially in contingency scenarios. The integration of renewable energy sources (RES), extreme weather, and cyber-attacks intensifies congestion management challenges and increases operators' cognitive load. Artificial Intelligence (AI), particularly reinforcement learning (RL) agents, shows great promise for supporting operators in real-time decision-making~\cite{Marot2021}. Traditionally reliant on experience and human intuition, congestion management now demands AI-human teaming. AI assistants can recommend real-time actions to operators, avoiding the need for full automation while addressing the growing complexity of power grid operations.



The European Union's AI Act emphasizes accuracy, robustness, and cybersecurity throughout AI lifecycles. Article 15 demands ``measures to prevent, detect, respond to, resolve and control for attacks trying to manipulate the training data set, or pre-trained components used in training, inputs designed to cause the AI model to make a mistake, confidentiality attacks, or model flaws''. Robustness and resilience are crucial concepts in this discussion. Robustness refers to a system's ability to maintain performance under natural or adversarial perturbations, either locally (sample-specific) or globally (deterministic guarantees)~\cite{ISO24029}. Resilience, by contrast, is the ability of an AI system to prepare for, adapt to, and recover rapidly from perturbations or unexpected changes~\cite{Terminology2023}, emphasizing recovery. Notably, resilience is often used in power systems in a different but analogous context, where AI systems are replaced by power systems.
The AI Act highlights robustness but lacks methodologies to quantify it or address resilience. Organizations like ISO, IEC, and IEEE are developing standards to fill this gap. ISO/IEC 24029-2 assesses robustness in AI systems, defining properties like stability and sensitivity~\cite{ISO24029}. However, it primarily applies to classical AI applications (e.g., classification, regression)~\cite{Garrido2023} and lacks performance metrics for these properties.

In the academic literature, the technical robustness of AI-based systems has mainly been focused on computer vision problems with ANN of different architectures and types. One example is abstract interpretation, which approximates a potentially infinite set of behaviors with a finite representation for feedforward and convolutional neural network layers~\cite{Singh2019}. According to a recent literature review by Ilahi et al.,~\cite{Ilahi2022}, the number of publications and methodologies that study the impact of adversarial attacks in deep learning algorithms that do not use images as inputs is low~\cite{Ilahi2022}. One of the few works on RL is~\cite{Behzadan2017}, which studies how different exploration methods can enhance the robustness and resilience of deep RL against both training-time and test-time attacks.


Given the power system's high-risk nature, methodologies have been proposed to evaluate the robustness of AI-based decision systems. Dobbe et al. studied safety risks in use cases like frequency regulation and distributed energy resource control, focusing on robustness to policy shifts and adherence to constraints~\cite{Dobbe2020}. Formal verification methods for black-box models like artificial neural networks (ANNs) were introduced to understand boundaries and identify adversarial examples in power system security tasks~\cite{Venzke2021}. Zheng et al. proposed a gradient-based approach to generate adversarial perturbations for network topology optimization, identifying critical moments for these attacks~\cite{Zheng2021}. Chen et al.~developed adversarial agents to distort RL outputs and steer system trajectories in voltage regulation and congestion management tasks~\cite{Chen2021}.


Mechanisms to improve AI robustness include adversarial training and data protection. Tian et al. proposed protecting high-impact meters (e.g., encryption) and adversarial training for ANN-based state estimation~\cite{Tian2022}, while adversarial training was applied to power quality recognition with convolutional neural networks~\cite{Tian2021}. However, all these works lack a formal definition for robustness and resilience, in particular, performance metrics to assess the different properties of these overarching concepts in an RL setting. To address this gap in the literature, the present paper proposes a novel methodology for quantitatively assessing the robustness and resilience of RL-based agents designed for real-time congestion management in electrical grids. The focus is testing time (i.e., focused on conformity assessment of pre-trained AI models as defined in the AI Act), but the metrics are also applicable during the training time of the AI models.


This paper is structured as follows: Section~\ref{sec:use_case} outlines the congestion management use case. Section~\ref{sec:assessement} introduces perturbation agents and metrics for assessing robustness and resilience. A case study in an AI-friendly digital environment is presented in Section~\ref{sec:case_study}. Conclusions are in Section~\ref{sec:conclusions}.

\section{Congestion Management Use Case}\label{sec:use_case}

To evaluate the robustness and resilience of AI systems, this work focuses on the congestion management use case -- a high-risk task where human operators are supported by an AI assistant system to manage power flows and mitigate network congestion~\cite{Marot2021}. Specifically: a) a human dispatcher monitors the situation of the network at a 5-min resolution and identifies the need for remedial actions (e.g., topological changes, generation re-dispatch), ideally with anticipation over a few hours-ahead horizon; the AI-based information processing functions act as a triggering mechanism for intervention or categorizes situations for human solving; b) for a chosen task, the human dispatcher receives a recommendation for remedial action from the AI model, with information on predicted effect and confidence; c) the human dispatcher accepts the recommendation, requests new information or explanations, or looks for a different action guided by an exploration agent or via manual simulation with power flow calculation tools.


While AI assistants are powerful tools for grid operations, addressing associated risks is crucial. For RL, adversarial attacks target the state space, reward function, action space, and model space. In congestion management, focus is placed on state space perturbations due to: a) reward functions and AI models being secured in cyber-safe IT systems, enabling swift restoration or retraining in secure environments; b) action space operations being linked to OT systems with high cyber-security standards and limited attacker knowledge of network topology reduces attack efficacy. However, data-driven models remain vulnerable to imperceptible input perturbations~\cite{GoodFellow2015}, while missing or erroneous data is common in real-world networks. On the other hand, data-driven models are often vulnerable to small imperceptible perturbations to the input data~\cite{GoodFellow2015}. Furthermore, missing or erroneous data can be common in real-world grids.

\section{Assessment Methodology}\label{sec:assessement}


\subsection{Perturbation Agents}\label{subsec:agent}

As discussed in Section~\ref{sec:use_case}, this work considers a perturbation agent that generates adversarial examples in the input space of the AI system. While the agent cannot directly modify the actual state of the network (environment), it can intercept the input of the AI system and alter data measurements in this input, thereby influencing the system's outputs and decision-making. Although such a perturbation agent can be employed during the AI agent's training phase, this study uses it to evaluate a pre-trained AI agent in an operational context. The environment used in this work is an AI-friendly digital simulation (see Section~\ref{sec:case_study} for more details) that fully replicates electrical grid operations and is also employed for training RL-based agents. 

Three types of perturbation agents are considered: one simulating natural adversarial perturbations and the other two representing intentional manipulations (e.g., cyber-attacks) designed to remain undetectable by human experts. The first of these intentional agents uses gradient estimation to create adversarial examples in every step of the simulated environment, and the second uses RL to find the best timing and type for each attack.

\subsubsection{Random perturbation agent (RPA)}
Introduces random perturbations to simulate potential failures in the Supervisory Control And Data Acquisition (SCADA) or state estimation system. It replicates natural adversarial disruptions to the system, such as missing measurements. At each step in the environment -- corresponding to each instance when the human operator monitors the system -- the AI agent can inject a new perturbation, as illustrated in Algorithm \ref{algo1:random PA}.

\begin{algorithm}[bt]\small
    \caption{Random perturbation agent rules}
    \label{algo1:random PA}
    \KwData{$p, PP, s^{gen}, s^{load}, s^{flow}, \sigma^{gen}, \sigma^{load}, \sigma^{flow}$}
    Apply perturbations in $PP$ to $s^{gen}, s^{load}$ and $s^{flow}$\;
    $u \gets Uniform(0, 1)$\;
    \uIf{$u < p$}
    {
        Randomly choose value to perturb $s_i^m$ where $m \in \{gen, load, flow\}$ and $i \in \{1, \dots, |s^m|\}$\;
        $u \gets Unif(0, 1)$\;
        \uIf{$u < 0.2$}
        {
            $s_i^m \gets 0$\;
        }
        \uElse
        {
            $r \gets logNormal(0, \sigma^m)$\;
            $s_i^m \gets s_i^m \times r$\;
        }
        $k \gets Geometric(\frac{1}{6})$\;
        Add new perturbation to $PP$ for $k$ steps\;
    }
    Update remaining steps for perturbations in $PP$\;
\end{algorithm}

As an input the RPA needs the probability of introducing a new perturbation $p$, the set of previous perturbations that should still be applied $PP$, the actual data on generation ($s^{gen}$), load ($s^{load}$) and flow over power lines ($s^{flow}$) and the standard deviation of the introduced perturbation for each data group $\sigma^{gen}, \sigma^{load}$ and $\sigma^{flow}$.

In lines 2-3 of Algorithm~\ref{algo1:random PA}, a new perturbation is introduced with probability $p$. One specific value is chosen to be perturbed, either mimicking a complete failure to measure a value (line 7) or an incorrect measurement (lines 9-10). For the latter case, the lognormal distribution is chosen to ensure that the sign of the value remains the same and because we assume that the measurement errors approximately follow a normal distribution. This perturbation is then applied in the next $k$ steps by adding it to $PP$, with an average length of 6 steps, equivalent to 30 minutes. The geometric distribution is chosen for $k$ due to the memoryless property of the distribution.

\subsubsection{Gradient estimation perturbation agent (GEPA)}
The GEPA creates an adversarial example to minimize the output of the optimal action in the policy of the AI agent during every step. The adversarial examples are created using the combination of gradient estimation and the projected gradient descent used in \cite{Chen2021}. This estimation is necessary because this work focuses on ``black-box'' attacks. It is performed using Eq.~\ref{eq1: grad estimate}, where $L(\textbf{s})$ represents the target function that the attacker aims to manipulate. Specifically, $L(\textbf{s})$ is defined as the value corresponding to the optimal action in the AI agent's policy, and the objective is to minimize this value, making the action less desirable. The vector $e_i$ in Eq.~\ref{eq1: grad estimate}~has 1 in the $i^{th}$ position and 0 everywhere else. 
\vspace{-0.05cm}
\begin{align}
    g_i(\textbf{s}) = \frac{\partial L(\textbf{s})}{\partial s_i} \approx & \frac{L(\textbf{s} + 0.01\textbf{e}_i) - L(\textbf{s} - 0.01\textbf{e}_i)}{0.02}
    \label{eq1: grad estimate}
\end{align}

These estimated gradients can be computed for each value of an observed state and are used in the projected gradient descent algorithm -- see Algorithm~\ref{algo2:gradient}. 

\begin{algorithm}[bt]\small
    \caption{Gradient estimation perturbation agent}
    \label{algo2:gradient}
    \label{algo3: intel PA}
    \KwData{$s, W, \zeta, \xi$}
    $s^{adv} \gets s$\;
    \For{$w=i,\dots,W$}
    {
        Compute gradients $g(s^{adv})$\;
        $s^{adv} \gets s^{adv} (1 + \zeta sign(g))$\;
        Clip $s^{adv}$ to keep it between $s (1 - \xi)$ and $s (1 + \xi)$\;
    }
\end{algorithm}

In Algorithm~\ref{algo2:gradient}, $s$ is the vector representation of the state that needs to be perturbed, and $s^{adv}$ is the adversarial example. Additionally, $W$ is the number of iterations in the projected gradient descent, $\zeta$ is the step size in each iteration, and $\xi$ is the maximum perturbation, which is set to 10\%. This value was also used by Zheng et al.~\cite{Zheng2021}, who stated that the AC state estimation is unable to detect perturbations up to this threshold. 

\subsubsection{RL-based perturbation agent (RLPA)}
It actively aims to change the behavior of the AI agent with as little perturbation as possible. This agent builds upon the work of Garcia et al.~\cite{garcia2020}, who used multi-objective RL to train an agent capable of maximizing the reduction in long-term rewards while minimizing the magnitude of perturbations. However, here the upper bound of 10\% is used again, which simplifies the algorithm into a single-objective RL. Algorithm \ref{algo2: intel PA} shows the simplified algorithm used to train the perturbation agent.

\begin{algorithm}[bt]\small
    \caption{RL-based perturbation agent}
    \label{algo2: intel PA}
    \KwData{$H, K, P, \alpha, \epsilon$}
    Initialize $Q(s, a)$ arbitrarily\;
    \For{$h=i,\dots,H$}
    {
        Initialize state $s$\;
        \While{$k<K$ and stopping criterion not met}
        {
            $u \gets Uniform(0, 1)$\;
            \uIf{$u < \epsilon$}
            {
                Pick perturbation $p$ randomly from $P$\;
            }
            \uElse
            {
                $p \gets argmax_{p' \in P}Q(s, p')$\;
            }
            Apply perturbation $p$ to get $s^{adv}$\;
            Let AI agent choose action $a^{adv}$ based on $s^{adv}$\;
            Take action $a^{adv}$ and go to next state $s'$ with reward $R$\;
            $p' \gets argmax_{\hat{p} \in P}Q(s', \hat{a})$\;
            $Q(s, p) \gets Q(s, p) + \alpha (R + \gamma Q(s', p') - Q(s, p))$\;
            $s \gets s'$
        }
    }
\end{algorithm}

In Algorithm \ref{algo2: intel PA}, $H$ is the number of episodes, and $K$ is the maximum number of steps in each episode. The set $P$ contains all possible actions the perturbation agent can take, which in this case are the possible perturbations, and $\alpha$ and $\epsilon$ are the learning and exploration rates. The regular and perturbed states, $s$ and $s^{adv}$, are again used, and $a^{adv}$ is the action taken by the AI agent based on $s^{adv}$. In each step, the perturbation agent either performs a random perturbation or the best one according to the $Q$-values (lines 5-9), which are commonly used in RL frameworks and, in this work, represent how much the action is expected to lower the performance of the AI agent. Based on this perturbation, the environment will then go to the next state, $s'$, with an immediate reward of $R$. To estimate the long-term value of $p$, the best perturbation that can be performed in $s'$ is chosen in line 11. Afterward, in line 12, the $Q$-value for the original state and the performed perturbation, $Q(s, a)$, is updated using $R$ and the $Q$-value of the best case perturbation in $s'$. 

The perturbation agent can~\textit{do nothing} or perform a perturbation, which can change one or more of the values in the observed state and replace them with zero or a very large number. The agent can also create an adversarial example to make the AI agent take a specific action. The gradients are again estimated using Eq. \ref{eq1: grad estimate}. However, due to the running time, the Fast Gradient Sign Method ~\cite{goodfellow2014fgsm} is used here instead of the projected gradient descent. Additionally, the adversarial example will no longer be created at every step, but only if the RLPA thinks it will result in the biggest performance drop for the AI agent. Since these adversarial examples should make the AI agent perform a certain action, $L(\textbf{s})$ is defined as the value corresponding to that action in the policy of the AI agent, and it should be maximized. Using the estimated gradients, the adversarial example can be computed using Eq.~\ref{eq2: adv exmpl} as long as the maximum amount of perturbation $\xi$ is known, again set to 10\%.
\vspace{-0.05cm}
\begin{align}
    \textbf{s}^{adv} = \textbf{s} + \eta sign(\textbf{g})
    \label{eq2: adv exmpl}
\end{align}

Since including all possible combinations in $P$ would result in an exponentially growing number of possible perturbations, a greedy action space reduction heuristic based on the Teacher in the curriculum agent~\cite{Lehna2023} is applied to select the most promising combinations of up to three values in an observation. The number of possible targets for the adversarial examples is also reduced by grouping similar actions together and only including one action per group as a target.

\subsection{Robustness Metrics}\label{subsec:robust}

The technical robustness is evaluated using a perturbation agent (see Section~\ref{subsec:agent}) and using different metrics introduced in this section.

The first metric measures the difference in total rewards between the unperturbed and perturbed AI systems. This metric measures if the AI system can perform at the same level when introducing perturbations and can be calculated using the formula from Eq.~\ref{eq_reward}. $R^u_k$ is the reward obtained in step $k$ by the AI agent when no perturbations are performed on the input, and $R^{p}_k$ is the reward with perturbations. 
\vspace{-0.05cm}
    \begin{align}\label{eq_reward}
        \sum_{k=0}^K R^u_k - \sum_{k=0}^K R^{p}_k
    \end{align}

Another factor that can be used to determine the robustness of an AI system is the range of change in the output under perturbations. In this case, the output would be the action recommended by the AI agent, and the range of change is measured using two approaches. 

The first approach is to assess whether a particular decision holds for input variation (e.g., noise, missing data) in the same context by counting the number of times the decision the AI system takes with perturbations is different compared to the one made unperturbed, as shown in Eq.~\ref{eq:stab1} using the indicator function $\mathbbm{1}_{cond}$, which is equal to 1 if the condition $cond$, e.g. $x > 0$ or $x \in A$, is true and 0 otherwise. The Eq. also uses $a_k^{adv}$ and $a_k$, corresponding to the action taken in step $k$ with and without the adversarial agent, respectively.
\vspace{-0.05cm}
    \begin{align}\label{eq:stab1}
        \sum_{k=0}^K \mathbbm{1}_{a_k^{adv} \neq a_k}
    \end{align}  

The second approach to measure the range of change in the output is to look at how similar or different the new action with perturbation is to the original one. To do this, a similarity score is assigned to each pair of actions, $a^1$ and $a^2$. This score consists of two parts, one that corresponds to the same changes (Eq. \ref{eq:similarity1}) and one that accounts for the changes to the same substation in the network (Eq. \ref{eq:similarity2}). In Eq.~\ref{eq:similarity1}, the set $c^a$ consists of the changes that are made by action $a$, such as setting the origin of power line 1 at bus bar 1, and $c^{a^1, a^2} = c^{a^1} \cap c^{a^2}$ is the set of changes that are made in both actions $a^1$ and $a^2$. Additionally, $\hat{c}^{a^1, a^2}$ is the set of changes that are almost the same, e.g., an action setting the extremity of line 4 at bus 2 and an action setting it at bus 1. In Eq.~\ref{eq:similarity2}, $v^a$ is the set of all substations affected by action $a$, and $v^{a^1, a^2}$ is the set of substation affected by both $a^1$ and $a^2$.
\vspace{-0.05cm}
    \begin{align}
        C^{a^1, a^2} = \frac{1}{2} \left(|c^{a^1, a^2}| + \frac{|\hat{c}^{a^1, a^2}|}{2}\right) \left( \frac{1}{|c^{a^1}|} + \frac{1}{|c^{a^2}|} \right)
        \label{eq:similarity1}
    \end{align}
    
    \begin{align}
        V^{a^1, a^2} = \frac{1}{2} \left( \frac{|v^{a^1, a^2}|}{|v^{a^1}|} + \frac{|v^{a^1, a^2}|}{|v^{a^2}|} \right)
        \label{eq:similarity2}
    \end{align}
    
    \begin{align}
        \frac{V^{a^1, a^2} + C^{a^1, a^2}}{2}
        \label{eq:similarity3}
    \end{align}

The metrics defined in Eq.~\ref{eq:stab1}--\ref{eq:similarity3} allow the verification of the \textit{stability} property, as specified in ISO/IEC 24029-2~\cite{ISO24029}. This property evaluates whether the system's output remains consistent despite variations in the input and whether its performance is maintained under such conditions. In this case, stability is assessed by comparing the output of a pre-trained AI agent under perturbation to its expected output in the absence of any perturbation.

Another metric is the number of steps in the environment before a grid failure occurs. Using the set of all legal states $S$ and the indicator function, it is computed with Eq.~\ref{eq:failure}.
\vspace{-0.05cm}
    \begin{align}\label{eq:failure}
        \sum_{k=0}^K \mathbbm{1}_{s_k \in S}
    \end{align}

This metric corresponds to the \textit{reachability} property outlined in ISO/IEC 24029-2~\cite{ISO24029}, as it assesses whether the AI agent can effectively prevent a grid failure state (e.g., a cascading event resulting in a blackout) when subjected to perturbations.

Even if the perturbations fail to reduce the obtained reward or trigger a grid failure, they may still lead the AI system to take unnecessary actions, often incurring additional costs. To address this, another important metric is the reward per action. This metric is formulated as Eq.~\ref{eq:raction}, the decision for an agent to~\textit{do nothing} is denoted by $a^\emptyset$; this means that $\mathbbm{1}_{a_k \neq a^\emptyset}=1$ if the AI agent takes any action in step $k$ and 0 if the agent~\textit{does nothing}.
\vspace{-0.05cm}
    \begin{align}\label{eq:raction}
        \frac{\sum_{k=0}^K R_k}{\sum_{k=0}^K \mathbbm{1}_{a_k \neq a^\emptyset}}
    \end{align}

Additionally, analyzing data points that serve as weak spots in the system -- those capable of significantly altering the system's output with minimal changes -- offers valuable insights. This is measured by calculating the proportion of instances where perturbing each input value results in a change in the action taken by the AI agent. However, since a perturbation is applied to every value in every step for the GEPA, a threshold is defined to determine whether a value is significantly perturbed. This threshold is computed using the mean, $\mu_i$, and standard deviation, $\sigma_i$, of all changes to the given value as $\mu_i \pm \sigma_i$. Based on this threshold, we can define $\mathbbm{1}_{s_i \text{ is perturbed}}$ as the indicator function indicating whether the $i^\text{th}$ observation value is significantly perturbed or not. The metric can then be computed using Eq.~\ref{eq:count}.
\vspace{-0.05cm}
    \begin{align}\label{eq:count}
        \frac{\sum_{k=0}^K \mathbbm{1}_{a_k^{adv} \neq a_k} \mathbbm{1}_{s_ik \text{ is perturbed}}}{\sum_{k=0}^K \mathbbm{1}_{s_ik \text{ is perturbed}}} & \quad ,  \forall s_i \in s
    \end{align}
    

\subsection{Resilience Metrics}\label{subsec:resil}

The quantification of resilience is strongly related to the magnitude and duration of reward function performance degradation compared to an unperturbed system in the same context.

The first metric is the area between the reward curves of the unperturbed and perturbed AI system from the episode where the perturbations are introduced, $h^{p}$. This metric can be approximated using the trapezoidal rule for numerical integration as in Eq.~\ref{eq:trap}, with $\Delta R_{hk} = R^{p}_{hk} - R^{u}_{hk}$.
\vspace{-0.05cm}    
{
    \small
    \begin{align}\label{eq:trap}
        \sum_{k=1}^K \int_{h^{p}}^H \Delta R_{hk} dh \approx \sum_{k=1}^K \frac{\Delta R_{Hk} + \Delta R_{h^{p}k}}{2} + \sum_{i=h^{p} + 1}^{H-1}\Delta R_{ik}
    \end{align}
}

\begin{figure}[hbt]
    \centering
    \includegraphics[width=0.75\linewidth]{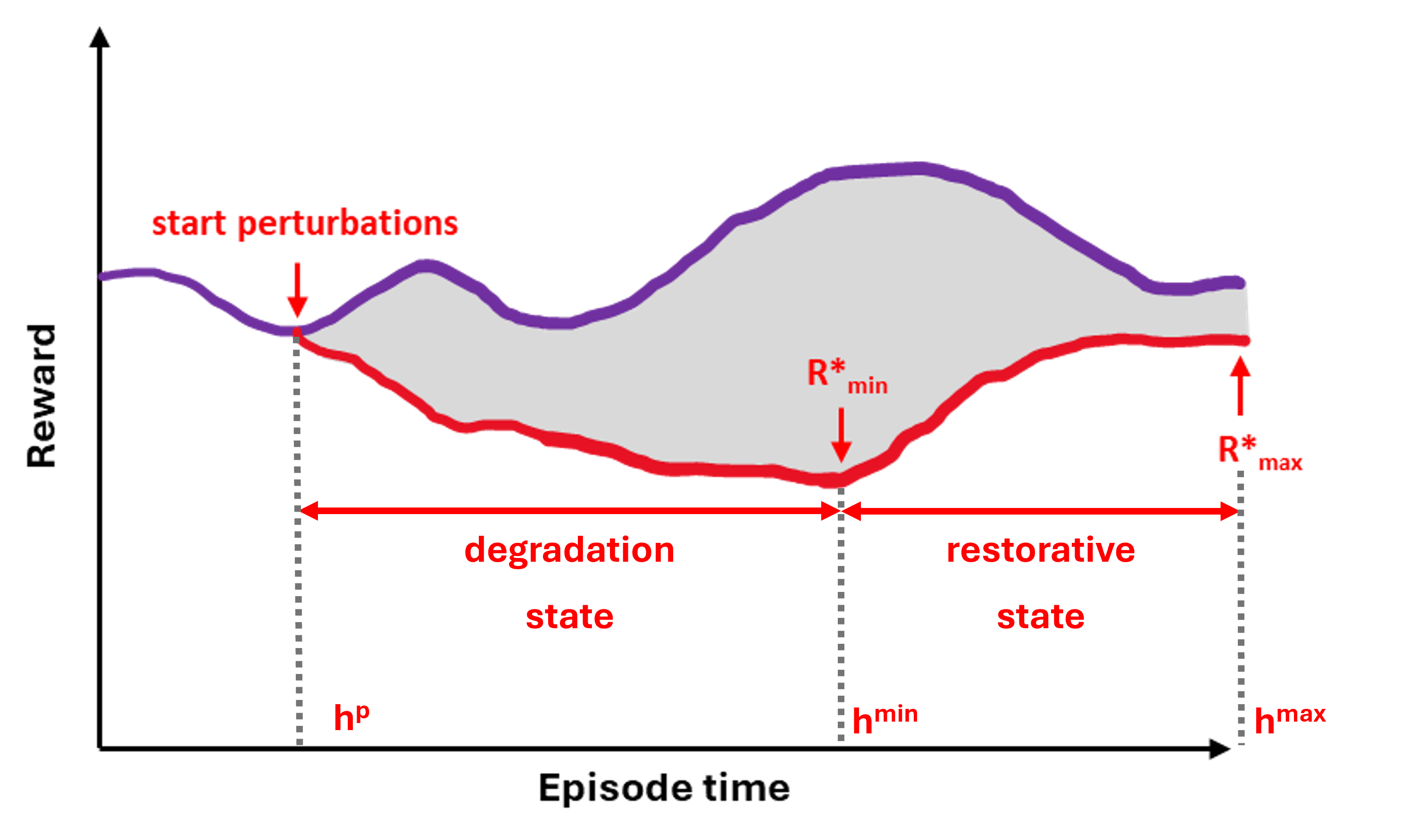}
    \caption{Degradation and restorative state during the testing of the AI system}
    \label{fig:resilience}
\end{figure}

The next metrics measure how quickly the AI system can adapt to the introduction of perturbations by counting how many episodes the degradation and restorative stages consist of. An example is depicted in Fig~\ref{fig:resilience}.

These metrics can be computed by first defining the episode with the lowest reward after perturbations, $h^{min}$, and the episode with the highest reward after $h^{min}$, $h^{max}$:
\vspace{-0.05cm}
    \begin{align}
        h^{min} = argmin_{h^{p}\leq h \leq H}\{\sum_{k=1}^K R_{hk}\}\\
        h^{max} = argmax_{h^{min} \leq h \leq H}\{\sum_{k=1}^K R_{hk}\} 
    \end{align}

Using these values, the degradation and restorative time can be calculated as $h^{min} - h^{p}$ and $h^{max} - h^{min}$, respectively.

Besides these metrics, it is valuable to examine the extent of performance deterioration and assess whether the system can recover to its original performance level without perturbations. To do this, the minimum reward in the degradation state and maximum reward in the restorative state can be computed with Eq.~\ref{eq:minmax}.
\vspace{-0.05cm}
    \begin{align}\label{eq:minmax}
        min_{h^{p}\leq h \leq H}\{\sum_{k=1}^K R_{hk}\} \\
        max_{h^{min} \leq h \leq H}\{\sum_{k=1}^K R_{hk}\}
    \end{align}

Finally, the similarity between the state of the grid with unperturbed and perturbed AI systems over time is used as a metric. It measures how drastically the actual state of the grid is affected after an action of the AI system is changed by perturbations and whether the AI system is able to revert these changes. In this work, the cosine similarity between the vector representations of the states, as seen in the equation below, is used as the metric, but other distance metrics could also be used (e.g., Euclidean).
\vspace{-0.05cm}
    \begin{align} \label{eq:cos_sim}
        \frac{\sum_{i=0}^{|s|} s^{adv}_i s_i}{\sqrt{\sum_{i=0}^{|s|}(s^{adv}_i)^2} \sqrt{\sum_{i=0}^{|s|}(s_i)^2}}
    \end{align}

Calculating this similarity at each step enables the identification of both degradation and recovery states in relation to the unperturbed state, similar to the approach used for the reward.

\section{Numerical Results}\label{sec:case_study}
\subsection{Case-study}

The analysis is conducted on the IEEE-14 bus system, which is a simple approximation of a power network and is available in the existing open-source AI-friendly digital environment called Grid2Op~\cite{Donnot2020}, developed for the Learning to Run a Power Network (L2RPN) competition series~\cite{Marot2021}. Grid2Op helps users develop both expert systems and RL-based topology controllers for power grid operation and control. It can be used to enable the development of an AI assistant in control centers for topology reconfigurations and re-dispatching, but in this work, it is used to evaluate resilience and robustness at test-time. 
The curriculum agent described in~\cite{Lehna2023}, and available as a baseline model in Grid2Op, is used as the AI agent to be analyzed in terms of robustness and resilience. The training of the RLPA is done using a Deep Q Networks training framework~\cite{mnih2013} built using PyTorch.

\subsection{Analysis}
To obtain the results, the average was taken over 35 episodes run in Grid2Op, each with a maximum number of 8064 steps. 

\subsubsection{Robustness}

\begin{figure}[hbt]
    \centering
    \includegraphics[width=\linewidth]{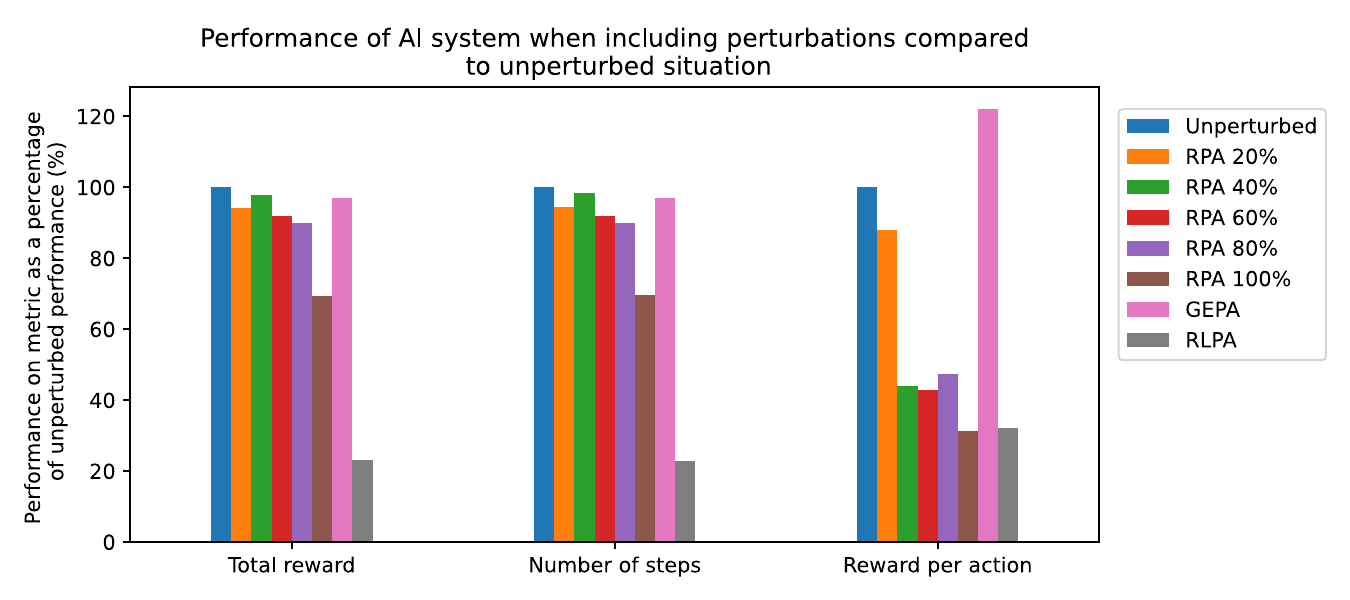}
    \caption{Robustness metrics as a percentage of the unperturbed performance of the AI system}
    \label{fig:metrics1}
\end{figure}

In Fig.~\ref{fig:metrics1}, the AI system's performance change on several robustness metrics is shown when including a perturbation agent. The values for each metric are scaled as a percentage of the performance of the system in an unperturbed environment, which is why the unperturbed values are all 100\%. For instance, the AI system's survival time with the RPA at $p = 20\%$ is approximately 95\%, meaning that in an environment where the system can operate for 1000 steps without perturbations, it would experience a complete grid failure after 950 steps when the perturbation agent is active.

From Fig.~\ref{fig:metrics1}, it is evident that the AI system performs well under both the RPA and GEPA, as the total reward and number of steps remain above 90\% in most cases, except for the RPA at $p=100\%$. However, for the RPA, the reward per action is significantly lower, indicating that the AI agent takes more actions to achieve the same reward level. In contrast, the reward per action for the GEPA exceeds 120\%, further demonstrating the AI agent's resilience to its perturbations. Conversely, the agent's performance under the RLPA is notably poor, with all three metrics below 30\%.

Table~\ref{tab:robust_metrics1} presents the robustness metrics that cannot be directly compared to the unperturbed performance. As Fig.~\ref{fig:metrics1} demonstrated significant variations in the number of steps per episode across different perturbation agents, the values have been normalized for consistency.

\begin{table}[hbt]
    \centering
    \caption{Robustness metrics that can not be compared to the unperturbed situation}
    \resizebox{\linewidth}{!}
    {
        \begin{tabular}{rrrrrrrr}
            \hline
             & \multicolumn{5}{c}{RPA} \\
             & 20\% & 40\% & 60\% & 80\% & 100\% & GEPA & RLPA \\
            \hline
            Actions changed & \multirow{2}{*}{6.092} & \multirow{2}{*}{14.819} & \multirow{2}{*}{19.283} & \multirow{2}{*}{26.636} & \multirow{2}{*}{33.745} & \multirow{2}{*}{1.665} & \multirow{2}{*}{53.414} \\
            per 1000 steps& & & & & & \\
            Similarity score & \multirow{2}{*}{0.005} & \multirow{2}{*}{0.005} & \multirow{2}{*}{0.005} & \multirow{2}{*}{0.004} & \multirow{2}{*}{0.004} & \multirow{2}{*}{0.240} & \multirow{2}{*}{0.025} \\
            per changed action & & & & & &\\

            \hline
        \end{tabular}
    }
    \label{tab:robust_metrics1}
\end{table}

It demonstrates that a higher probability of introducing perturbations increases the number of altered actions, while the alternatives chosen by the AI system in response to these changes become less similar to the optimal action. The GEPA cannot change as many actions and the alternative is quite similar. However, the RLPA is again able to drastically affect the performance of the AI agent, as it causes the most action changes, and while the similarity score is higher than for the RPA, it is still a lot lower than for the GEPA.

\begin{figure}[bt]
    \centering
    \includegraphics[width=\linewidth]{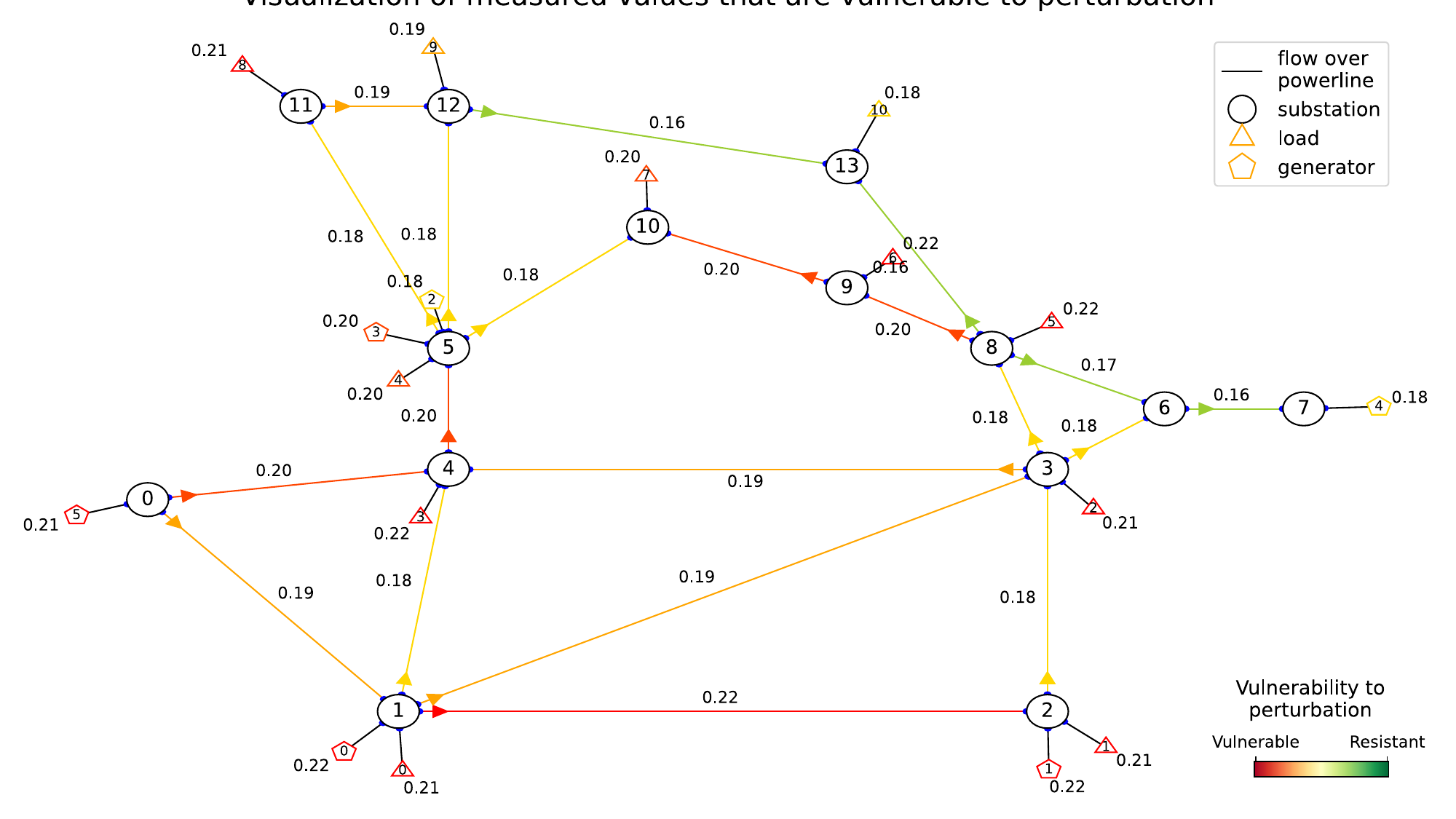}
    \caption{Visualization of values in the grid that are vulnerable to perturbation}
    \label{fig:metrics2}
\end{figure}

Fig.~\ref{fig:metrics2} depicts the power grid areas vulnerable to perturbations using Eq.~\ref{eq:count} for the GEPA. A traffic light color scheme indicates vulnerability: green for less likely, red for more likely, and yellow in between. Observations include generation, load, and power line flow values. It shows that power lines and the load at substation 9 are vulnerable to perturbations, suggesting potential volatility.

\subsubsection{Resilience}

\begin{figure}[hbt]
    \centering
    \includegraphics[width=0.9\linewidth]{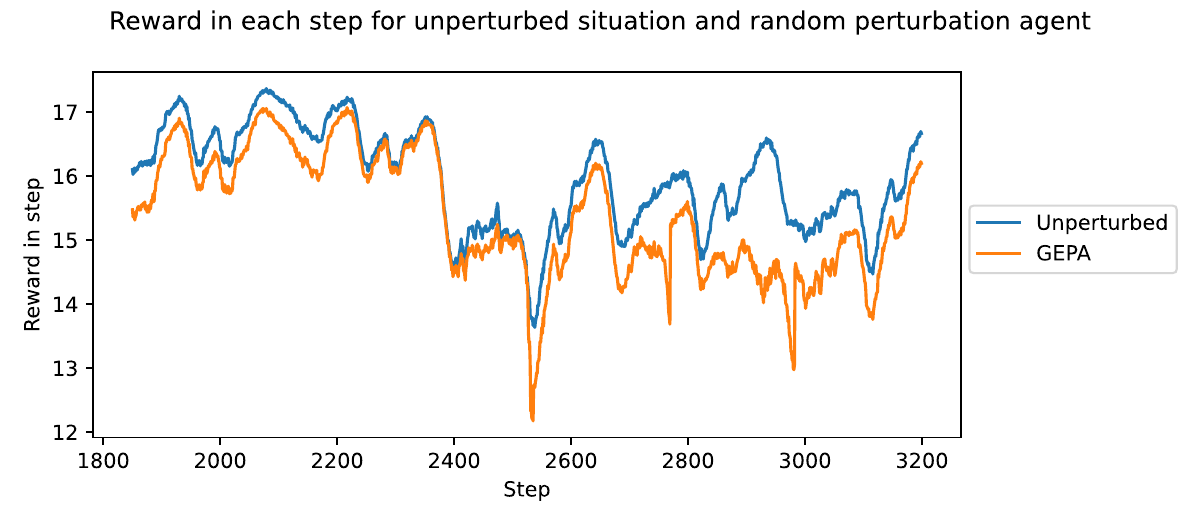}
    \caption{Example of the reward obtained in each step between step 2450 and 3200 of an episode in an environment with the GEPA}
    \label{fig:metrics3}
\end{figure}

Fig.~\ref{fig:metrics3} depicts the reward obtained in each step during several steps of an episode in the testing phase for the environment with the GEPA.  

This figure is included to illustrate the ability of the AI agent to recover after a drop-off in performance caused by perturbations. Between steps 2800 and 3100, a relatively large difference can be seen between the unperturbed and perturbed reward, making it easier to see the degradation and restorative state of the AI agent. From Fig.~\ref{fig:metrics3}, it can be observed that $h_{min} \approx 2990$ and $h_{max} \approx 3100$, indicating that the degradation state lasts for approximately 190 steps, while the restorative state persists for around 110 steps. It can also be seen that there can be multiple degradation states in a single episode as another degradation and restoration can be seen between steps 2500 and 2800. For this reason, the average number of degradations in an episode is also reported. Table~\ref{tab:res_metrics1} presents the average performance of the AI agent on these metrics in environments with different perturbation agents.   

\begin{table}[hbt]
    \centering
    \caption{Resilience metrics for the reward obtained during an episode in the testing phase}
    \label{tab:res_metrics1}
    \resizebox{\linewidth}{!}
    {
        \begin{tabular}{rrrrrrrr}
            \hline
             & \multicolumn{5}{c}{RPA} \\
             & 20\% & 40\% & 60\% & 80\% & 100\% & GEPA & RLPA \\
            \hline
            Degr. time & 478.81 & 435.50 & 452.90 & 480.26 & 430.86 & 476.07 & 424.22 \\
            Rest. time & 527.12 & 480.62 & 411.23 & 392.96 & 404.83 & 467.15 & 517.52 \\
            $max(\Delta R)$ & 9.50 & 10.76 & 9.95 & 10.93 & 10.64 & 8.13 & 8.75 \\
            $min(\Delta R)$ & -8.53 & -7.20 & -8.27 & -8.19 & -8.23 & -8.54 & -6.47 \\
            \# degr. per & \multirow{2}{*}{0.64} & \multirow{2}{*}{0.80} & \multirow{2}{*}{0.88} & \multirow{2}{*}{0.84} & \multirow{2}{*}{0.84} & \multirow{2}{*}{0.62} & \multirow{2}{*}{0.38} \\
            1000 steps & & & & & & \\
            Area per & \multirow{2}{*}{47.44} & \multirow{2}{*}{145.91} & \multirow{2}{*}{55.08} & \multirow{2}{*}{111.44} & \multirow{2}{*}{77.96} & \multirow{2}{*}{9.75} & \multirow{2}{*}{109.34} \\
            1000 steps & & & & & & \\
            \hline
        \end{tabular}
    }
\end{table}


The results indicate that the GEPA leads to longer degradation and restorative periods than the RPAs. However, the minimum reward achieved is not as low, and the maximum reward is higher. This suggests that the AI agent is relatively effective at withstanding and recovering from the perturbations introduced by the GEPA. The maximum reward is negative because the perturbed reward is higher than the unperturbed reward in some steps, as the AI agent tries to recover part of the lost reward. It can also be seen that the number of degradations per episode is lower, which might be caused by the fact that each degradation takes longer. The area between the unperturbed and perturbed reward curve is on the lower end.

\begin{figure}[hbt]
    \centering
    \includegraphics[width=0.9\linewidth]{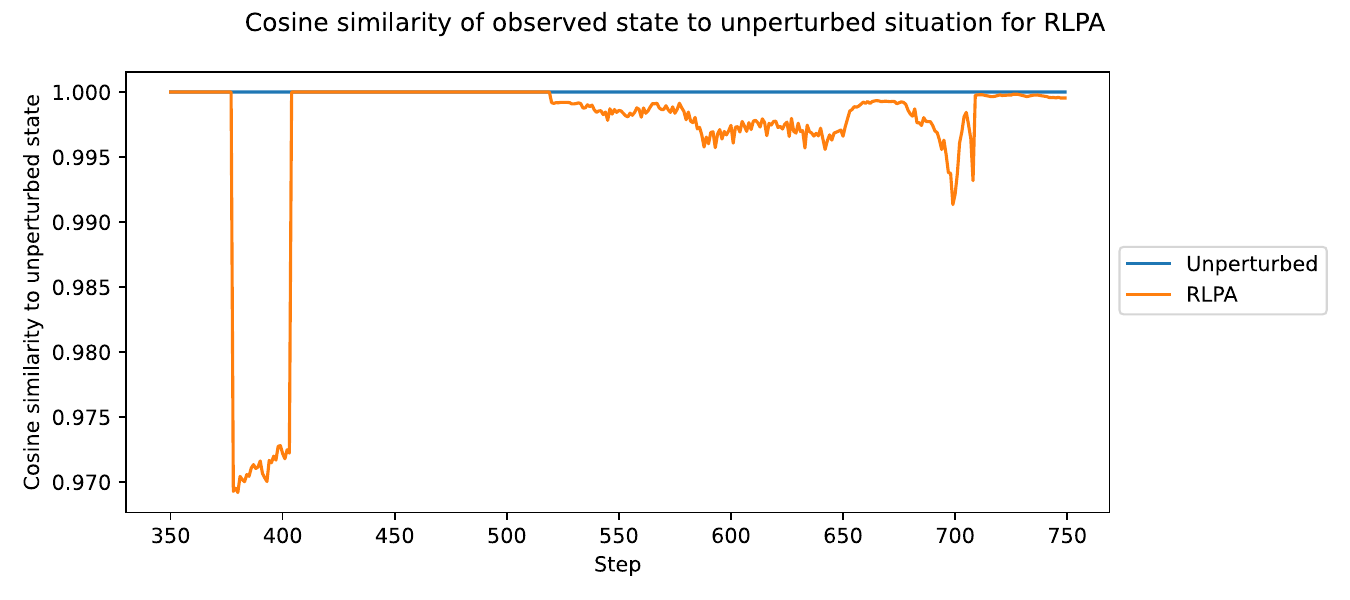}
    \caption{Example of cosine similarity to unperturbed state in each step between steps 350 and 750 of an episode in the environment with the RLPA}
    \label{fig:metrics4}
\end{figure}

Fig.~\ref{fig:metrics4} depicts the cosine similarity, as defined in Eq.~\ref{eq:cos_sim}, at each step of an episode for both the unperturbed environment and the environment with the RLPA for a subset of the steps. It can be seen that the AI agent can recover the state to the unperturbed state perfectly in about 20 steps after the first drop-off around step 380, even though the drop-off is quite big. Although the second degradation after 510 steps is smaller, the agent seems to struggle a bit more with it, but after 200 steps, it is able to recover to a score very close to 1. 

The averages per episode for the metrics on the cosine similarity can be seen in Table~\ref{tab:res_metrics2}. For the state similarity to the unperturbed state, the GEPA still results in quite long restorative times, although the degradation time is not as long in this case. The lowest point after perturbation is again quite reasonable, and since the maximum reward is close to zero, the agent is also able to recover well.  

\begin{table}[hbt]
    \centering
    \caption{Resilience metrics for the similarity of the state to the unperturbed state}
    \label{tab:res_metrics2}
    \resizebox{\linewidth}{!}
    {
        \begin{tabular}{rrrrrrrr}
            \hline
             & \multicolumn{5}{c}{RPA} \\
             & 20\% & 40\% & 60\% & 80\% & 100\% & GEPA & RLPA \\
            \hline
            Degr. time & 256.73 & 257.11 & 201.98 & 196.66 & 190.20 & 261.34 & 169.02 \\
            Rest. time & 331.32 & 339.43 & 321.02 & 287.94 & 304.46 & 451.92 & 1057.03 \\
            $max(\Delta R)$ & 3.31 & 3.27 & 3.38 & 3.62 & 3.78 & 3.21 & 3.75 \\
            $min(\Delta R)$ & 0.95 & 0.87 & 0.91 & 1.06 & 1.23 & 0.79 & 1.09 \\
            \# degr. per & \multirow{2}{*}{1.00} & \multirow{2}{*}{1.20} & \multirow{2}{*}{1.45} & \multirow{2}{*}{1.55} & \multirow{2}{*}{1.79} & \multirow{2}{*}{0.76} & \multirow{2}{*}{1.04} \\
            1000 steps& & & & & & \\
            Area per & \multirow{2}{*}{7.62} & \multirow{2}{*}{9.83} & \multirow{2}{*}{11.45} & \multirow{2}{*}{11.97} & \multirow{2}{*}{13.45} & \multirow{2}{*}{6.44} & \multirow{2}{*}{10.25} \\
            1000 steps& & & & & & \\

            \hline
        \end{tabular}
    }
\end{table}

The metrics described in Sections~\ref{subsec:robust} and \ref{subsec:resil} are computed and show that RL-based algorithms can have a significant variation in performance (i.e., reward function) but are capable of recovering up to a point, provided the congestion problem is solved. 

\section{Conclusions}\label{sec:conclusions}

This work bridges the gap between ongoing standardization efforts and practical implementation in quantifying robustness and resilience in mission-critical tasks, offering a methodology and metrics to evaluate AI agents under natural and adversarial perturbation scenarios quantitatively. A case study in the Grid2Op environment showed that the proposed approach effectively identifies vulnerabilities in AI decision-making systems and varying susceptibility to perturbations. RL-based perturbation agents reveal significant weaknesses in AI performance compared to random or gradient-based perturbations. 

Future work consists of a) developing more intelligent adversarial agents, e.g., based on an RL algorithm, that includes features (e.g., where to attack, perturbation budget) in the reward function, and b) developing a methodology to co-identify thresholds (with the decision-maker) for the metrics proposed in this paper and determine if a certain AI-based system is certified for safe operation.
\vspace{-0.3cm}

\bibliographystyle{IEEEtran}
\bibliography{references}

\end{document}